# PERFORMANCE ANALYSIS OF NEURO GENETIC ALGORITHM APPLIED ON DETECTING PROPORTION OF COMPONENTS IN MANHOLE GAS MIXTURE

Varun Kumar Ojha[1], Paramartha Dutta[1], and Hiranmay Saha[2]

[1]Department of Computer & System Sciences, Visva Bharati University, Santiniketan, West Bengal, India
`varun.kumar.ojha@gmail.com, Paramartha.dutta@gmail.com`
[2]Center of Excellence for Green Energy and Sensor Systems, Bengal Engineering & Science University, Shibpur, West Bengal, India
`sahahiran@gmail.com.com`

## ABSTRACT

*The article presents performance analysis of a real valued neuro genetic algorithm applied for the detection of proportion of the gases found in manhole gas mixture. The neural network (NN) trained using genetic algorithm (GA) leads to concept of neuro genetic algorithm, which is used for implementing an intelligent sensory system for the detection of component gases present in manhole gas mixture Usually a manhole gas mixture contains several toxic gases like Hydrogen Sulfide, Ammonia, Methane, Carbon Dioxide, Nitrogen Oxide, and Carbon Monoxide. A semiconductor based gas sensor array used for sensing manhole gas components is an integral part of the proposed intelligent system. It consists of many sensor elements, where each sensor element is responsible for sensing particular gas component. Multiple sensors of different gases used for detecting gas mixture of multiple gases, results in cross-sensitivity. The cross-sensitivity is a major issue and the problem is viewed as pattern recognition problem. The objective of this article is to present performance analysis of the real valued neuro genetic algorithm which is applied for multiple gas detection.*

## KEYWORDS

*Gas mixture, Gas sensor array, Cross-sensitivity, Floating point, Genetic algorithm, Neural network, Optimization, Computational complexity*

## 1. INTRODUCTION

The environmental awareness and pollution control are major concern today. This article offers detail study about the implementation mechanisms and performance analysis of neuro genetic algorithm for the development of an intelligent sensory system for detection of proportion of different component gases present in typical manhole gas mixture. When the domestic and industrial waste products are decomposes into the sewer pipeline network, poisonous gaseous mixture is formed, known as manhole gas mixture. This gaseous mixture usually contains gases like, Hydrogen Sulfide ($H_2S$), Ammonia ($NH_3$), Methane ($CH_4$), Carbon Dioxide ($CO_2$), Nitrogen Oxide ($NO_x$), etc [1], [2], [5], [8]. The manholes are built across the sewer pipeline network for cleaning and maintenance purpose. A person as conventional practice has to get down into the pipeline network to serve this purpose. Many pedestrian also becomes vulnerable with these manholes as because the manholes are built on the roads or on the road sides. In recent days few instances of death including municipality labourers and pedestrians are reported due to toxic gas exposures [10]. There is also an environmental pollution concern due to exposed manhole gases. The twin problems lead us to mould our research involvements in this direction.

We are using neural network classifier to design an intelligent sensory system for the detection of proportion of component gases present in manhole gas mixture. The neural network has to be developed such that it can act like an intelligent agent who can report the proportional presence of toxic gas components in the manholes when it dropped into manholes for the detection. In present article the gas detection problem is treated as a pattern recognition problem, where the neural network classifier [16] is trained in supervised mode and we are using the genetic algorithm for training of neural network. As the training of the neural network is done externally using the genetic algorithm so it given rise to a hybrid approach of neuro genetic algorithm. Such intelligent sensory systems may help labourers to be watchful against the presence of toxic gases before working into the manholes. In [11], [12], [13], [14], and [15] concerned authors presented their respective approaches towards solution to manhole gas detection issue. A semiconductor based gas sensor array containing distinct semiconductor type gas sensors sense the presence of gases according to concentration in manhole gas mixture. Sensed values by the sensor array are cross-sensitive because multiple gases are present in the manhole gas mixture and even if the distinct gas sensors are sensitive to their target gases they are influenced by other gases too. Our objective is to train the neural network such that the cross-sensitivity effect can be minimized and the proposed system can generate flawless report.

In this article, first we discuss the mechanisms in designing of a gas detection system and the section includes the idea of the complete design of the experimental setup for gas detection. Then we discuss the process of the formation of data samples, which are used for experimentation. In this section we included discussion on the cross-sensitivity issue. Then we have defined the training pattern used for the training of neural network. In neuro genetic technique section we elaborately discussed the hybrid concept of neuro genetic algorithm and we also discuss the genetic algorithm. The performance analysis section is the central theme of this article. This section also provides rich discussion on the performance analysis of the neuro genetic algorithm. The performance analysis is performed based on the tuning different parameters of genetic algorithm. Finally the result section presents clear picture about the output representation of the intelligent sensory system.

## 2. MECHANISM

The mechanism section deals with the procedure of the developing an experimental setup for the detection of mixed gases. Here we present an overview of the gas detection system followed by the data formation technique and definition of training pattern for the neural network classifier.

### 2.1. Overview of the Gas Detection System

The basic model of intelligent sensory system for detection of manhole gas mixture is shown in the Figure 1. In [3], [6], [18] concerned authors have demonstrated their ideas and perspectives toward the construction of a gas detection system and they have shown their own mechanisms for detecting gas components.

The developed gas detection system constitutes of three modules, input module, intelligent module and output module. The input module is an integration of gas chamber, gas sensor array and data pre-processing block. The intelligent module receive data from the input module and after performing the computation on those data it sends its result to the output module for presenting system output in more structured way. The sample of gas mixture is collected into a gas mixture chamber and subsequently allowed to pass over the semiconductor based gas sensor array [6], [7], [9]. The sensor elements present in the gas sensor array are responsive to their target gas. Although the gas sensors elements are made to detect their target gases they are showing sensitivity towards other gases too. So, the sensor array response is always involving cross-sensitivity effect [7]. The pre-processing block receives sensed data values from the gas sensor array and in this received data values are normalized before feeding to the neural network classifier. The neural network has to be trained using the normalized data values. The output module does the task of denormalization of the data coming out of the neural network. The output module also generates alarm if any of the toxic gas components exceeds their safety limit. For the training of the network, several data samples are produced. The section subsequent to this describes the data formation or collection process.

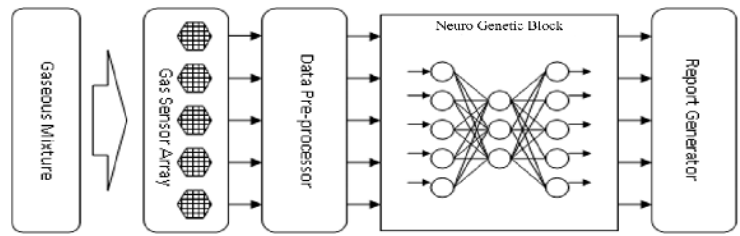

Figure 1.  Overview of Gas Detection System

The Figure is showing the flow of the work or the order of the task which may handle during a typical gas detection process. At first gas mixture chamber collect gases and passed it over gas sensor array. The pre-processor block let us, to provide normalized data value to neural network. The neural network performs its computation on normalized data and sends it to report generator module

## 2.2. Semiconductor based Gas Sensor Array

The metal oxide semiconductor gas sensors are used to form gas sensor array. N number of distinct sensor element of n gases constitutes a one dimensional gas sensor array. The MOS sensors are basically resistance type electrical sensors. A resistance type sensor respond as change in resistance on change in the concentration of gases. The change in resistance is given as $R_s/R_0$ (change ratio of sensor resistance) where the $R_s$ are the change in resistance of the MOS sensor and the $R_0$ is the base resistance value. A typical arrangement of a gas sensor array is shown in Figure 2. The circuitry shown in Figure 2 is developed in the laboratory at Bengal Engineering and Science University.

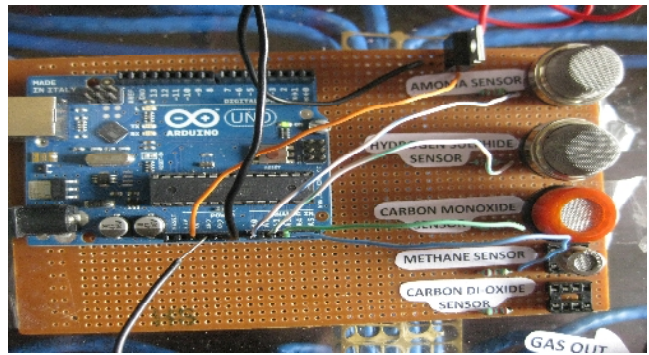

Figure 2.  Semiconductor based Gas Sensor Array

## 2.2. Data Collection

Initial step in data sample formation is the collection of information about the safety limits of the component gases found in manhole gas mixture. Then we prepare several gas mixture samples by mixing gas components in different combination of concentration. The concentrations of the gases in the mixture are taken around their safety limits. The known gas mixture is a synthetic mixture of gases in the known concentration. To prepare a data sample, the mixture of known concentration of gases is blown over the sensor array and the sensor responses are collected in tabular form. In this way several data samples are prepared. A typical example of such data sample is shown in the Table 1. Focusing to the first and second row of the Table 1 we can appreciate the inherent cross-sensitivity effect in the responses of sensors. From the first and second sample it is observed that the concentration of only Methane gas is increased and even if there is changes in the concentration of only Methane gas, there is change occur in responses of all the sensors including the sensor for Methane itself. It is indicating that the prepared data sample is containing the cross-sensitive effect and it is also observed that this cross-sensitivity effect is not random rather it is following some characteristics and pattern. So the sensor responses of the gas sensor array may not be used directly to predict or to report the concentration of the gases in manhole gas mixture. To predict or forecast the concentration of the gases in the manhole gas mixture we need to use intelligent system with pattern recognition techniques.

The data sample shown in Table 1 is prepared using mixture of five gases. The Table 1 is having three major columns for sample number, for sample gas mixture and for sensor responses respectively. All the samples are mixed in different combination of concentration value of gases shown in second column. The values of sensor response when sample mixtures are blown over gas sensor array in reset condition of sensor array are shown in third column.

Table 1.  The Prepared Data Sample

| # Sample | Sample mixture (ppm value) | | | | | Sensor response | | | | |
|---|---|---|---|---|---|---|---|---|---|---|
| | $NH_3$ | CO | $H_2S$ | $NO_2$ | $CH_4$ | $NH_3$ | CO | $H_2S$ | $NO_2$ | $CH_4$ |
| 1 | 50 | 100 | 100 | 100 | 2000 | 0.0531 | 0.0863 | 0.0733 | 0.0267 | 0.1101 |
| 2 | 50 | 100 | 100 | 100 | 5000 | 0.0812 | 0.1122 | 0.0749 | 0.0333 | 0.1934 |
| 3 | 50 | 100 | 100 | 200 | 2000 | 0.0963 | 0.1182 | 0.0929 | 0.0577 | 0.1195 |
| 4 | 50 | 100 | 200 | 200 | 5000 | 0.1212 | 0.12905 | 0.1291 | 0.0699 | 0.2086 |
| 5 | 50 | 100 | 200 | 400 | 2000 | 0.1451 | 0.1495 | 0.1399 | 0.0795 | 0.1207 |
| 6 | 100 | 200 | 200 | 400 | 5000 | 0.1569 | 0.1573 | 0.1526 | 0.0799 | 0.2559 |
| 7 | 100 | 200 | 100 | 200 | 2000 | 0.1693 | 0.1699 | 0.0722 | 0.0615 | 0.1400 |
| 8 | 100 | 200 | 100 | 200 | 5000 | 0.1715 | 0.1798 | 0.0883 | 0.0705 | 0.3000 |
| 9 | 100 | 200 | 100 | 400 | 2000 | 0.1821 | 0.2231 | 0.0996 | 0.1302 | 0.1544 |
| 10 | 100 | 200 | 200 | 400 | 5000 | 0.1924 | 0.2584 | 0.1869 | 0.1648 | 0.3124 |

## 2.3. Data Pre-processing

With raw data neural network may not forecast accurately so the normalization of the data sample is necessary before feeding it to the neural network. By normalization we mean to accumulate data within range 0 and 1. The standardization of input data for the neural network makes its training faster and reduces the chance of getting stuck in local optima. It also reduces the adversity which may come during the testing of the neural network. The pre-processing block normalizes the data samples according to equations (1) and (2). According to equation (1) the $NC_{si}$ (normalized concentration) of gas $H_2S$ of sample 2 is given by 100/5000 where the 100 appearing in the numerator is the $C_{si}$ (concentration of the gas itself) value and the 5000 in the denominator is the $C_{max}$ (maximum concentration among all the samples) value. Similarly, the sensor responses are also normalized according to equation (2) where, $NR_{si}$ is normalized sensor response, $R_{si}$ is sensors individual response and $R_{max}$ is the maximum response among all the samples.

$$NC_{si} = C_{si} \, / \, C_{max} \qquad\qquad\qquad\qquad (1)$$

$$NR_{si} = R_{si} \, / \, R_{max} \qquad\qquad\qquad\qquad (2)$$

The normalized values shown in the Table 2 are used for training of the neural network. Once the network is trained it is ready to be operative in test environment. The output of the neural network is need to be denormalized to report the system output in terms of concentration (parts per million) of gases present in the given test environment. The denormalization is simple method where the value of network output is multiplied with $C_{max}$ (maximum concentration among all the samples) value. The denormalization method is given in equation (3)

$$\text{System Output} = \text{Network Output} \quad C_{max} \qquad\qquad\qquad (3)$$

The Table 2 contains the normalized value corresponding to the Table 1. The Table 2 is having three major columns, for sample number, second for sample gas mixture and third for sensor responses respectively. The second column contains the normalized value of the concentration of gas mixture. The third column contains the normalized value of sensor responses.

Table 2. The normalized data sample.

| # Sample | Sample mixture | | | | | Sensor response | | | | |
|---|---|---|---|---|---|---|---|---|---|---|
| | $NH_3$ | CO | $H_2S$ | $NO_2$ | $CH_4$ | $NH_3$ | CO | $H_2S$ | $NO_2$ | $CH_4$ |
| 1 | 0.01 | 0.02 | 0.02 | 0.02 | 0.4 | 0.1699 | 0.2762 | 0.2346 | 0.0854 | 0.3524 |
| 2 | 0.01 | 0.02 | 0.02 | 0.02 | 1.0 | 0.2599 | 0.3591 | 0.2397 | 0.1065 | 0.6190 |
| 3 | 0.01 | 0.02 | 0.02 | 0.04 | 0.4 | 0.3082 | 0.3783 | 0.2973 | 0.1846 | 0.3825 |
| 4 | 0.01 | 0.02 | 0.04 | 0.04 | 1.0 | 0.3879 | 0.4130 | 0.4132 | 0.2237 | 0.6677 |
| 5 | 0.01 | 0.02 | 0.04 | 0.08 | 0.4 | 0.4644 | 0.4785 | 0.4478 | 0.2544 | 0.3863 |
| 6 | 0.02 | 0.04 | 0.04 | 0.08 | 1.0 | 0.5022 | 0.5035 | 0.4884 | 0.2557 | 0.8191 |
| 7 | 0.02 | 0.04 | 0.02 | 0.04 | 0.4 | 0.5419 | 0.5438 | 0.2311 | 0.1968 | 0.4483 |
| 8 | 0.02 | 0.04 | 0.02 | 0.04 | 1.0 | 0.5489 | 0.5755 | 0.282 | 0.2256 | 0.9603 |
| 9 | 0.02 | 0.04 | 0.02 | 0.08 | 0.4 | 0.5829 | 0.7141 | 0.3188 | 0.4167 | 0.4942 |
| 10 | 0.02 | 0.04 | 0.04 | 0.08 | 1.0 | 0.6158 | 0.8271 | 0.5982 | 0.5275 | 1.0000 |

## 2.4. Neural Network Training Pattern

The supervised mode of training is used for the training of neural network classifier. So the training pattern should constitute of input vector and target vector. It is also mentioned above that the normalized sensor responses are given as input to the neural network. So the input vector 'I' is created using the normalized values of the sensor responses. In the given data sample input vector is a five element vector. One vector element is for each gas in the sample gas mixture. For sample 1 of Table 2, the input vector 'I' can be represented as follows.

I = [0.1699, 0.2762, 0.2346, 0.0854, 0.3524]

The output intelligent system is presented in terms of the concentration of gases. So the target vector 'T' is formed using the gas mixture sample. In the given data sample target vector is a five element vector. One vector element is for each gas in the sample gas mixture. For sample 1 of Table 2, the target vector 'T' can be represented as follows.

T = [0.01, 0.02, 0.02, 0.02, 0.4]

From the Table 2 it can be observed that the second major column is taken as target vector and third major column is taken as input vector.

## 3. NEURO GENETIC TECHNIQUE

Genetic algorithm is search algorithm based on the mechanics of natural selection and natural genetics [4]. In Neuro Genetic approach, the genetic algorithm searches out optimized synaptic weights for the neural network [19]. In [13] authors have shown the use of neuro genetic algorithm in gas detection system.

### 3.1. The Neuro Genetic Algorithm

We are using the real valued genetic algorithm for training of the neural network. The soft computing tools such as neural network and genetic algorithm are coupled in such a manner that it forms a concept of neuro genetic algorithm [19]. A flow chart of the neuro genetic algorithm is shown in the Figure 3 is offering a minimization problem where it tries to minimize the sum squared error induced by the neural network. The objective of finding out the minimum sum squared error is done by searching optimal synaptic weights of the neural network. The synaptic weights in the neural network are real valued (float value) so the genetic algorithm operates on the real value (float value) [20], [21]. The chromosome in the genetic algorithm is created using float values (real values). The steps of genetic algorithm are described in the following section.

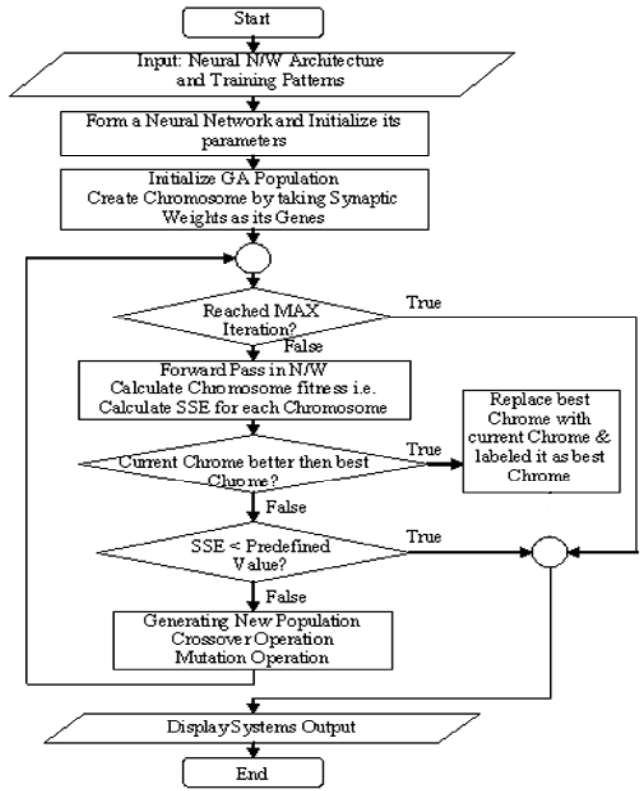

Figure 3. A Schematic of Neuro Genetic Algorithm

### 3.1.1. The Chromosome Structure

The chromosomes in the neuro genetic algorithm are created using the synaptic weights of the neural network [19]. The synaptic weights in the neural network are float value so the chromosome is also created by the combination of float values. In neuro genetic algorithm the synaptic weights are considered as genes of the chromosome. Each gene (a single float value) is represented (or encoded) using 32 bit IEEE 754 floating point format (shown in Figure 4(b)). If a neural network is having total of N synaptic weights then a chromosome has N number of genes. And total length of the chromosome is 32 x N number of bits. Synaptic weights encoding into chromosome stricture is shown in the Figure 4(a).

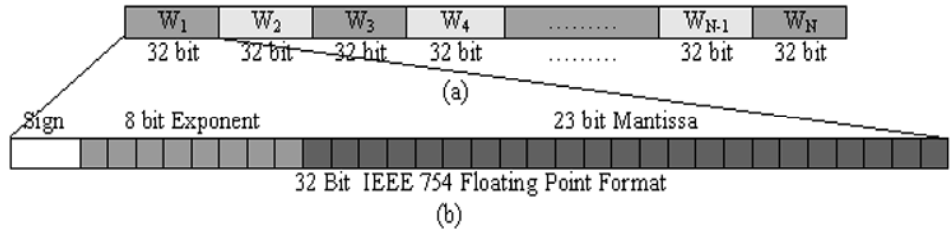

Figure 4. (a) Encoding Synaptic Weights into Chromosomes (b) The Sketch of Single Gene ($W_i$) in Neuro Genetic Algorithm.

The synaptic weights $W_1$, $W_2$, $W_3$, $W_4$, $W_{N-1}$, $W_N$ are the float values and its corresponding 32 bit binary representation is taken as a gene into the chromosome. And these genes will again decode into float value using IEEE 754 floating point formats according to equation (4) [20].

Float Value = -1S x Base$^{E-127}$ x 1.M          (4)

### 3.1.2. Selection Operation

For chromosome selection, the fitness proportionate selection (FPS) has been used. In the fitness proportionate selection method the fitness values of each chromosome are normalized. To normalize fitness value of each chromosome we have to divide fitness of each chromosome by the summation of all fitness value. The population is sorted in descending order of the fitness values of the chromosome using simple sorting algorithm. Then accumulated normalized fitness values for each individual are computed by computing sum of its own fitness and fitness of all the previous individuals. To select an individual we generate a random number R between 0 and 1 and the first individual whose accumulated normalized value greater than R is selected.

### 3.1.3. Composite Single Point Crossover Operation

The crossover operation defined here is a composite single point crossover. The composite single point crossover is different from multipoint crossover. In the multipoint crossover operation the crossover is performed at multiple points into the entire chromosome and the genes are exchanged within these selected points. In the composite single point crossover, chromosome is uniformly fragmented into N parts where N is the total number of genes in the chromosome. The crossover operation is performed only within the similar positioned fragment of two chromosomes at a random chosen single index point. There is a uniform crossover operation probability (p) typically p = 0.8 is set for crossover operation in each fragments. A pictorial representation of crossover operation is provided in Figure 5(a).

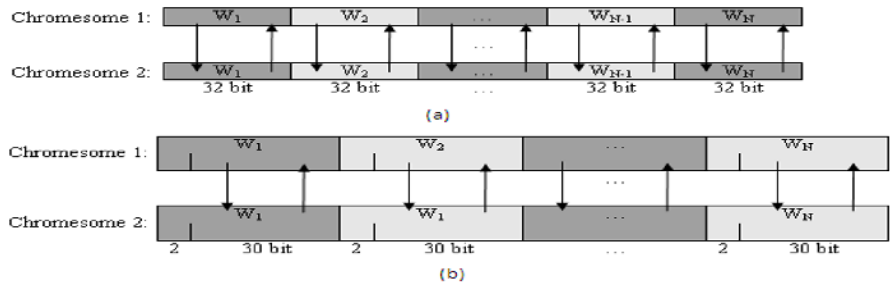

Figure 5. (a) Composite Single Point Crossover in Neuro Genetic Algorithm (b) Restrictive Composite Single Point Crossover in Neuro Genetic Algorithm

While doing the crossover operation we face a NAN (not a number) error problem. The value NaN (Not a Number) is used to represent a value that does not represent a real number [20], [21]. To stay away from the NaN setback we opted for a restrictive crossover mechanism. In this technique crossover is restrict to be made only within the last 29 bits or 30 bits (LSB) of a gene or in other words first 3 bits or 2 bits (MSB) are not taking part in the crossover operation. A pictorial representation of this restrictive crossover is provided in Figure 5(b).

### 3.1.4. Composite Single Point Mutation Operation

The Composite Single Point Mutation is an operation which flips a randomly chosen bit into each gene of a chromosome. Here mutation is performed with low probability pm = 0.02. Similar to the crossover operation the mutation is also performed within only last 29 bits or 30 bits to avoid NAN problem

### 3.1.5. Computing Fitness

The Genetic Algorithm task is to perform minimization or maximization of an objective function. In our Neuro Genetic Algorithm, the objective is the minimization of sum squared error induced by the neural network while the network is undergone into a training process. In [17] the sum squared error is defined as it is provided in equation (5). As already mentioned a chromosome accommodates all the synaptic weights of a neural network. The chromosomes actually decoded as synaptic weights for the neural network and upon this synaptic weights SSE are evaluated thus evaluated SSE become the fitness value of those chromosomes.

$$\text{SSE} = 1/2 \quad (O_{pi} - t_{pi})^2 \quad \text{for all p \& i} \tag{5}$$

Where, $O_{pi}$ and $t_{pi}$ are the actual and desired outputs respectively realized at the output layer, 'p' is the input pattern vector and 'i' is the number of nodes in the output layer.

### 3.1.6. Terminating Criteria

The algorithm terminates either when the sum of squared error reached to an acceptable minimum or when the algorithm completes its maximum allowed iteration

## 4. PERFORMANCE ANALYSIS

The neuro genetic algorithm is implemented and trained using the data sample. Thereafter the performance of the neuro genetic algorithm is analyzed. The performance of any algorithm is watched in terms of the quality of its result and the computational cost pied to obtain that result.

By doing the performance analysis we mean to observe the strength and weaknesses of neuro genetic algorithm. We are presenting the performance analysis of the neuro genetic algorithm with respect to its different parameter. In this section we also discuss about the time and space complexity met by the neuro genetic algorithm.

## 4.1. Parameters Tuning

The parameters that have been chosen for the performance analysis of the neuro genetic algorithm are neural network configuration, population size (number of chromosomes), number of generations (iterations), crossover rate ($p_c$), mutation rate ($p_m$), and initial search space (solution space). We have analyzed the quality of the result/output, i.e. observing the average of sum of squared error taken over ten instances run of the neuro genetic algorithm.

### 4.1.1. Average Sum of Squared Error vs. Neural Network Configuration

Initially we want to find out optimum neural network configuration. So we need to observe the quality of the result obtained against the different neural network configuration. Keeping other parameters fixed to certain values, we change the number of nodes at hidden layer in the neural network to draw the performance characteristic of the algorithm. We set genetic algorithm population size to hundred, number of generation to thousand, crossover rate to 0.7, mutation rate to 0.05 and initial search space range to [–1.5 to 1.5]. The Figure 6 exhibit the performance of neuro genetic algorithm based on network configuration, where the x-axis indicates the change in the number of nodes at hidden layer (H) in (5 – H – 5) network architecture and the y-axis indicates the average value of the sum of squared error obtained against different network configuration. It is already mentioned that the data set prepared using the sample mixture which is containing five gases, so we need to use five nodes at input layer and five nodes at output layer. Analysis based on network configuration is kind of structural training of the neural network. From the Figure 6 it clearly derived that algorithm performs better around the network configuration 5 – 3 – 5 and 5 – 4 – 5 where value 3 and 4 indicates the number of nodes at hidden layer and for the entire configuration higher than 5 – 3 – 5 and 5 – 4 – 5 the performance of the algorithm is becoming poor and poor.

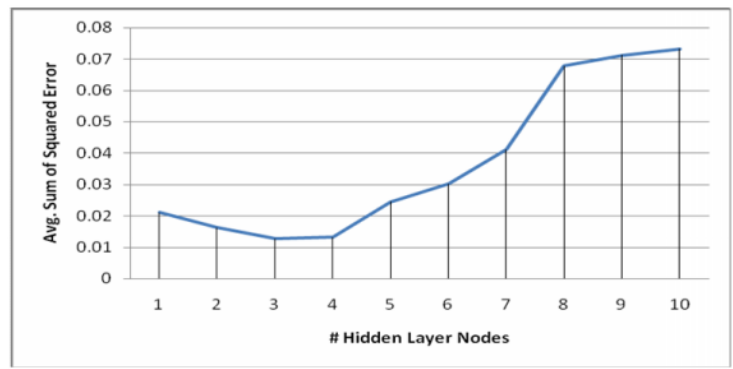

Figure 6. Average SSE vs. Neural Network Configuration

### 4.1.2. Average Sum of Squared Error vs. Population Size

After the structural training of the network we need to establish the population size for the algorithm. Keeping other parameters fixed, we tune the population size between size 10 and 100 to draw the performance characteristic of the algorithm based on population size. We set genetic algorithm population size to hundred, number of generation to thousand, crossover rate to 0.7, mutation rate to 0.05 and initial search space range to [−1.5 to 1.5]. The Figure 5 exhibit the performance of neuro genetic algorithm based on population size, where the x-axis indicates the change in the population size and the y-axis indicates the average value of the sum of squared error obtained against the different population size taken. From the Figure 7 it is clearly visible that for the increasing values of the population sizes the performance of algorithm improves. At the population size equal to hundred the error get down from 50% to below 10% and it is approximately around 5%. So we set the population size equal to hundred chromosomes for analyzing performance of algorithm for other parameters.

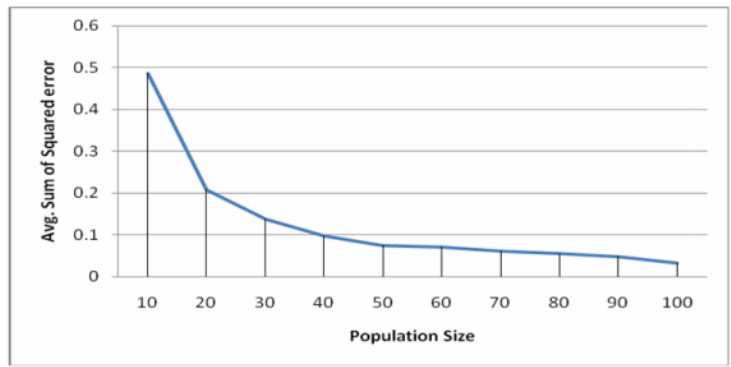

Figure 7. Average SSE vs. Population Size

### 4.1.3. Average Sum of Squared Error vs. Number of Generations

The Figure 8 exhibits the performance of genetic algorithm against different number of generations. Here the average sum of squared error is computed against the differ iteration. We set genetic algorithm population size to hundred, number of generation to thousand, crossover rate to 0.7, mutation rate to 0.05 and initial search space range to [−1.5 to 1.5]. From the graph vide Figure 8 we found that the sum of squared error decreases for all the increasing values of generations. It is also observed from the graph that at iteration 1000 the error value get down below 3 percent.

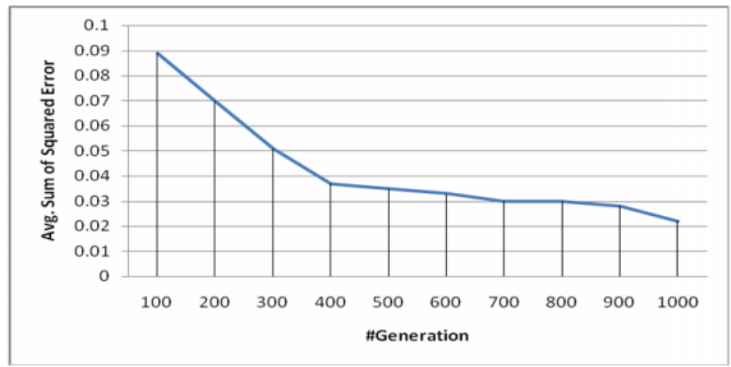

Figure 8. Average SSE vs. Number of Generations

### 4.1.4. Average Sum of Squared Error vs. Crossover Rate

The crossover rate plays important role in genetic algorithm. If the crossover probability $p_c$ is equal to 0 then none of the selected chromosomes takes part in crossover to produce new generation and the each selected chromosomes are copied into the new generation. If the crossover probability $p_c$ is equal to 1 then all of the selected chromosomes takes part in crossover to produce new generation and none of the chromosome can transit or survive in the new generation only offspring are available in new generation. None of the mentioned case is advantageous so we need to have an optimum crossover probability to obtained fittest generation. Keeping other parameters fixed, we vary $p_c$ from 0.5 to 1 to draw the performance characteristic of the algorithm based on the crossover probability. We set genetic algorithm population size to hundred, network configuration to $5 - 3 - 5$, number of generation to thousand, mutation rate to 0.05 and initial search space range to [–1.5 to 1.5]. The Figure 9 exhibits the performance of neuro genetic algorithm based on crossover rate, where x-axis indicates the change in the probability of crossover and y-axis indicates the average value of the sum of squared error obtained against different probability of crossover. It is found that initially the performance of algorithm improves for the increasing values of crossover probability but increasing its values after $p_c = 0.8$ the performance of the algorithm is started falling.

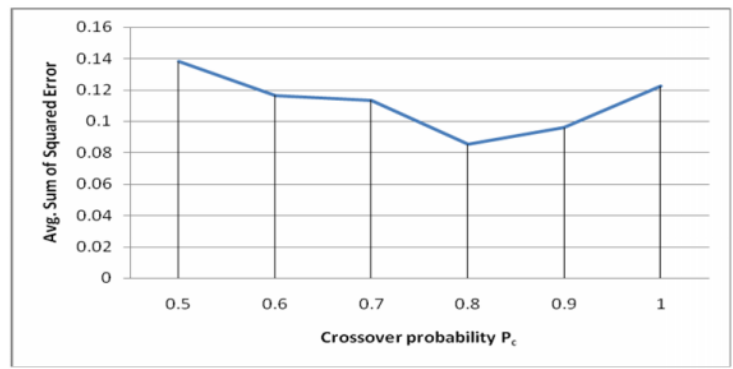

Figure 9.  Average SSE vs. Crossover probability

### 4.1.5. Average Sum of Squared Error vs. Mutation Rate

The mutation operation also plays important role in genetic algorithm. Due to mutation a part of chromosome changed. If the mutation probability pm is equal to 0 then none of the selected chromosomes takes part in mutation and the each selected chromosomes are copied into the new generation as it is. If the mutation probability pm is equal to 1 then all of the selected chromosomes takes part in mutation. Here none the case is advantageous so we need to have an optimum mutation probability to obtained fittest generation. Keeping other parameters fixed, we vary $p_m$ from 0.01 to 0.05 to draw the performance characteristic of the algorithm based on the mutation probability. We set genetic algorithm population size to hundred, network configuration to $5 - 3 - 5$, number of generation to thousand, crossover rate to 0.8 and initial search space range to [–1.5 to 1.5]. The Figure 10 exhibits the performance of neuro genetic algorithm based on mutation rate, where x-axis indicates the change in the probability of mutation and y-axis indicates the average value of the sum of squared error obtained against different probability of crossover. It is found that initially the performance of algorithm improves for the increasing values of crossover probability but increasing its values after $p_m = 0.03$ the performance of the algorithm is started falling.

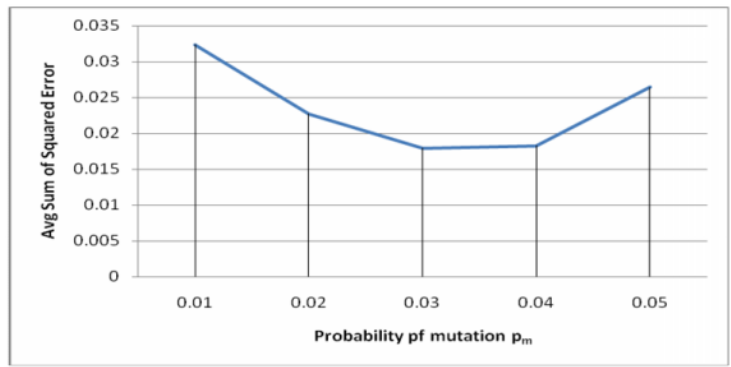

Figure 10. Average SSE vs. Mutation probability

### 4.1.6. Average Sum of Squared Error vs. Search Space

After analyzing the performance of the algorithm for the parameters population size, network configuration, crossover rate, mutation rate it become necessary to check the performance of the algorithm against the range of search space, because at some extant the performance of the neural network depends on the initial choice of synaptic weights. The Figure 11 exhibits the performance of neuro genetic algorithm based on initial choice solution space range, where x-axis indicates the change in the choice of range for initial solution and y-axis indicates the average value of the sum of squared error obtained against different range of initial solution. Analyzing the graph vide Figure 11 it is derived that the performance of algorithm is coming down on widening the search space means the average sum of squared error continue increases each time the range widen. Here value 0.5 along x-axis represents range of the search space and it is between [– 0.5 and + 0.5]. It is observed that average SSE increases when the range of initial search space is widening.

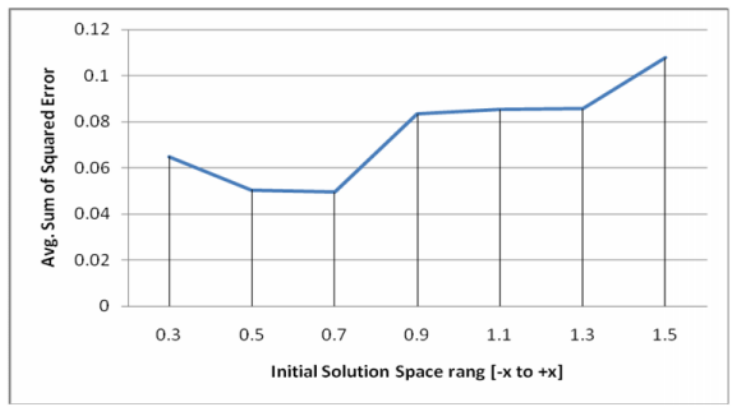

Figure 11. Average SSE vs. Initial Solution Space

### 4.2. Complexity Analysis

The complexity analysis of any algorithm includes computation complexity (time complexity) analysis and space complexity analysis. The subsequent sections offer time and space complexity analysis of real valued neuro genetic algorithm.

### 4.2.1. Analysis of Computational Complexity

The computation complexity of the neuro genetic algorithm is same as the computational complexity of genetic algorithm. In genetic algorithm, at first fitness values of each individuals of the initial population are computed and based on these fitness values individuals are selected to form new generation using the crossover and mutation operations. The basic operations performed in genetic algorithm are fitness calculation, selection operation, crossover operation and mutation operation. In [22] the author discuss about the time and space complexity met by a genetic algorithm. The computational complexity of the real valued genetic algorithm depends on the cost of evaluating the fitness value of the chromosomes and the cost of producing a new generation. So the computational complexity of neuro genetic algorithm entirely depends on the number of times fitness function evaluated and new generation produced. The fitness value evaluation for the neuro genetic algorithm is nothing but the computation of sum of squared error value induced by a neural network for given inputs. The production of new population includes three operations, selection operation, crossover operation and mutation operation. Let E denotes time complexity of real valued neuro genetic algorithm. So the time complexity of neuro genetic algorithm E is given by the expression (5).

$$E = t \quad m \quad (E_f + E_g) \tag{5}$$

Where, t is the numbers of iteration, m is the total population size, $E_f$ is the cost of evaluating a single fitness function and $E_g$ is the cost of producing new generation. Let the cost of fitness value evaluation is $E_f$ is equal to the cost of computing sum of squared error. From the expression (4) it is clear that to compute sum of squared error we have to compute square errors obtained at output layer for each input pattern P. The computation of error at output layer is nothing but two dimensional matrix multiplications. The cost of computing two dimensional matrix multiplications is $O(n^2)$. So the cost of computing sum of squared error is $E_f = P \quad O(n^2)$ where P is the total number of patterns. The cost of producing new generation is the summation of the cost met by the selection operation, crossover operation and mutation operation. The cost of selection operation is equal to cost of sorting (selection sort takes $O(n)$ in average case) and normalization of fitness value( to normalize we need cost of $O(n)$). If pc is the probability of crossover, $p_m$ is the probability of mutation, and n is the size of problem encode into chromosome so, the cost need for a composite single point crossover operation is $O(\ p_c \quad n)$ and the cost of computing composite single point mutation operation is $O(p_m \quad n)$. Hence the time complexity of neuro genetic algorithm stands to $O(t \quad m \quad n^2)$. Due to the computation cost of genetic algorithm depends on the crossover probability and mutation probability the time complexity genetic algorithm is a probabilistic cost rather than deterministic cost.

### 4.2.1. Analysis of Space Complexity

The space complexity of the neuro genetic algorithm is due to storage of chromosomes in memory. Let the length of chromosome is n and the size of total population is m so the space complexity of the neuro genetic algorithm is of the $O(m \quad n)$.

## 5. RESULTS REPRESENTATION

From the prepared data samples we consider 80% samples for training and 20% for testing purpose. The output of the network is denormalized according to equation (3) to report systems output in terms of the concentration of the gas component present in the gaseous mixture. Here we are providing a sample test result obtained for the input sample 2 which is given in Table 2. Where, the input vector is [0.260, 0.346, 0.240, 0.142, 0.843] and target vector is [0.01, 0.02, 0.02, 0.02, 1.0]. Finally we are producing this input vector for giving input to the neural network for testing purpose. The test result of the neural network for this input vector is shown in Table 3.

Table 3. Systems Results Presentation

| Testing i/p or i/p node no. | Normalized i/p value (Sample 2 of Table 2) | n/w Actual o/p | System's o/p in PPM |
|---|---|---|---|
| 1 | 0.260 | 0.016 | 0080 |
| 2 | 0.346 | 0.022 | 0110 |
| 3 | 0.240 | 0.023 | 0115 |
| 4 | 0.142 | 0.022 | 0110 |
| 5 | 0.843 | 0.993 | 4965 |

The first column represents the node number at input layer. The values shown in second column are normalized value of sensors response of gas mixture sample 2 of Table 2.The third column represents the output which is directly obtained from the neural network and its denormalized value is shown in the fourth column.

## 6. CONCLUSIONS

In this article, we discuss mechanisms for the development of an intelligent sensory system comprising of semiconductor based gas sensor array and neural network classifier. The article offers solution to the manhole gas detection problem. The gas detection problem is traded as a pattern recognition problem. We have shown that how a neural network classifier is trained using genetic algorithm. The article provides detail study about the implementation of neuro genetic algorithm. The article emphasise the significance cross-sensitivity issue. The cross-sensitivity which is contained by the prepared data sample is filtered out during the training of the neural network classifier. The neural network has been trained such that it can forecast actual concentration level of the gas components present in manhole gas mixture. In the performance analysis section we have provided the analysis of the performance of neuro genetic algorithm based on the tuning of different parameters of the genetic algorithm. The article provides the calculation of computational expensiveness of neuro genetic algorithm. Finally we present clear idea to present the systems output in terms of the concentration of gases present in manhole gas mixture.

## ACKNOWLEDGEMENTS

Authors wish to acknowledge Department of Science & Technology (Govt. of India) for the financial supports vide Project No.: IDP/IND/02/2009 to carry out this research.

## Authors

Mr. Varun Kumar Ojha did his Bachelor of Technology in Computer science & Engineering from West Bengal University of Technology in the year 2008 and did his Master of Technology in Computer Science & Engineering from Kalyani Government Engineering College, West Bengal University of Technology in the year 2011. He is working as fulltime Research Fellow in DST Govt. of India funded project at Visva-Bharati University. He has published about 7 research articles. He is a Member of Associated Computing Machinery (ACM) 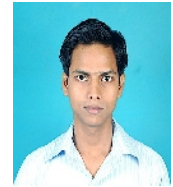

Prof. (Dr.) Paramartha Dutta did his Bachelors and Masters in Statistics from the Indian Statistical Institute, Calcutta in the years 1988 and 1990 respectively. He afterwards completed his Master of Technology in Computer science from the same Institute in the year 1993 and Doctor of Philosophy in Engineering from the BESU, Shibpur in 2005. He has served in the capacity of research personnel in various projects funded by Govt. of India, which include DRDO, CSIR, Indian Statistical Institute, Calcutta etc. Dr. Dutta is now a Professor in the Department of Computer and system Sciences of the Visva Bharati University, West Bengal, India. He is a Member of Associated Computing Machinery (ACM), IEEE, Computer Society, USA 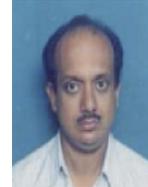

Prof. Hiranmay Saha received his M. Tech degree in Radiophysics and Electronics from University of Calcutta in 1967 and Ph. D degree in Solar Cells and Systems from Jadavpur University in 1977. He was former Chairman of Solar Energy Division (Eastern Region), the Ministry of New and Renewable Energy, Government of India and Advisor of WBREDA (Dept. of Power and N.E.S., Govt. of West Bengal). He is associated with Jadavpur University, as Professor in the Department of Electronics and Telecommunication Engineering. Currently he is the Professor-in-charge of 'the CGESS', BESUS, WB. He is Fellow, IEE (UK), IETE and Member IEEE, NCS (WB), Dept. of Science & Technology (WB), WBREDA. 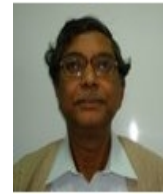